\documentclass[conference]{IEEEtran}

\usepackage{cite}
\usepackage{amsmath, amssymb}
\usepackage{graphicx}
\usepackage{hyperref}
\usepackage{newtxtext, url}

\begin{document}

\title{How AI Fails: An Interactive Pedagogical Tool for Demonstrating Dialectal Bias in Automated Toxicity Models}

\author{
\IEEEauthorblockN{Subhojit Ghimire}
\IEEEauthorblockA{
Independent Researcher\\
Gorkha, Nepal\\
\texttt{SubhojitGhimire@proton.me}
}
}

\maketitle

\begin{abstract}
Now that AI-driven moderation has become pervasive in everyday life, we often hear claims that ``the AI is biased.'' While this is often said jokingly, the light-hearted remark reflects a deeper concern. How can we be certain that an online post flagged as ``inappropriate'' was not simply the victim of a biased algorithm?

This paper investigates this problem using a dual approach. First, I conduct a quantitative benchmark of a widely used toxicity model (unitary/toxic-bert) to measure performance disparity between text in African-American English (AAE) and Standard American English (SAE). The benchmark reveals a clear, systematic bias: on average, the model scores AAE text as 1.8 times more toxic and 8.8 times higher for ``identity hate.''

Second, I introduce an interactive pedagogical tool that makes these abstract biases tangible. The tool’s core mechanic, a user-controlled sensitivity threshold, demonstrates that the biased score itself is not the only harm; instead, the more concerning harm is the human-set, seemingly neutral policy that ultimately operationalises discrimination. This work provides both statistical evidence of disparate impact and a public-facing tool designed to foster critical AI literacy.
\end{abstract}


\section{Introduction}

Online content moderation tools extensively use automated toxicity models to evaluate content and decide whether to flag it as inappropriate \cite{gongane2022}. This practice is not new, but it is becoming more prevalent today with the introduction of AI, where automated tools are now publicised as AI systems. Even though there is nothing particularly intelligent about these systems, which is why I prefer calling them just ``algorithms'', the AI namesake has managed to pique the public's interest. Even the general population has begun using and interacting with these ``AI''-moderation tools.

In one sense, this is a good thing. If more people are aware of a certain tool, more people will try to understand its intricacies and begin to question and dissect it. This public auditing, in itself, can ensure that any hidden malpractice is brought to light and that appropriate actions can be taken to correct it, preventing further harm. This is especially crucial in systems that operate as ``black boxes,'' raising concerns about their fairness and real-world impact. When an AI model's decision-making is opaque, it becomes difficult to challenge or even question its outcomes \cite{doi:10.1177/2053951715622512}.

The arbitrary ``black box'' problem should not be dismissed as a mere technical inconvenience, especially when the stakes in a critical society are high. If a model tasked with detecting ``toxic'' language contains hidden, systematic biases, it will not simply produce random errors but may systematically silence the voices of minority groups \cite{Noble+2018}. An AI model is trained on a vast dataset, so it learns to associate linguistic features and dialects with the demographic groups represented in that dataset \cite{blodgett-etal-2016-demographic}. But is the dataset used to train the model fair? Does it contain equal representation of dialects across demographics? Or is it biased towards one majority group? If so, the result is a disparate impact, wherein a system, under the guise of safety and moderation, disproportionately flags benign text from one community while ignoring harmful text from another \cite{10.1145/3630106.3659036}.

Many thorough studies have been conducted on this topic, and while some expert tools exist to visualise this problem, they are neither readily available nor designed as pedagogical instruments for a non-expert audience. This has created a significant gap between expert findings and public understanding. It is one thing to read a statistical report on bias; it is another to see and interact with that bias in real time. This paper aims to bridge that gap.

I present a dual-pronged contribution to investigate and demonstrate this problem. First, I conduct a quantitative audit of a foundational, widely used toxicity model \textit{unitary/toxic-bert} \cite{toxicbert} to scientifically measure its performance disparity on African-American English (AAE) versus Standard American English (SAE). Then, I introduce a novel interactive pedagogical tool that I developed, designed explicitly to make the bias tangible. This tool displays scores for various sentiment categories after analysing the sentiment of an input text and allows users to adjust a sensitivity threshold slider. This interaction is designed to provide a key insight, as users realise that this slider represents not a neutral, model-determined boundary, but a human-set policy, which ultimately leads to discriminatory outcomes.

\section{Related Work}

My research is situated at the intersection of two fields: 1. The quantitative study of algorithmic bias in NLP models and, 2. the Human-Computer Interaction (HCI) challenge of making AI understandable to the public. This section reviews work in both areas to establish the context for my contribution.

Blodgett et al., in their foundational study, demonstrated that language identification models often misidentify AAE \cite{blodgett-etal-2016-demographic}. They provided a large-scale Twitter corpus, the same one I use in this paper, to help researchers investigate these disparities \cite{twitteraae}. Their work showed that demographic dialectal variation is a significant confounding variable for many NLP tasks. Building on this, Sap et al. analysed several large-scale datasets used for hate speech detection \cite{sap-etal-2019-risk}. They presented a critical finding that all of them contained a strong, spurious correlation between AAE linguistic features and ``toxicity'', meaning that models trained on this data learned to associate AAE with abusive language, even when the context was benign. Parikh et al. further established that this systematic, data-driven bias is not a minor flaw but a significant, measurable problem in widely-deployed systems, such as in the medical field, where it can perpetuate health disparities \cite{11515928}.

Amidst the challenge of bridging the gap between expert and public understanding, a series of research has emerged focused on building tools for AI literacy. These tools can be broadly categorised into two parts: diagnostic tools for experts and pedagogical tools for the public. Tenney et al.’s Language Interpretability Tool (LIT) is a powerful example of a diagnostic tool for experts, allowing for deep analysis of model internals \cite{tenney-etal-2020-language}. Similarly, Biaslyze: The NLP Bias Identification Toolkit is a powerful Python package that offers a concrete entry point for impact assessment within NLP models, along with mitigation measures \cite{biaslyze}. Kabir et al.’s STILE and Viswanath et al.’s FairPy are two other noteworthy toolkits aimed at measuring and mitigating biases in language models \cite{10.1145/3613904.3642111,viswanath2025fairpytoolkitevaluationprediction}. While these tools serve as invaluable analysers for experts, they are not designed for a non-technical audience. They are ``expert-facing'' and require a high degree of technical knowledge to interpret.

Raz et al., on the other hand, developed \textit{Face Mis-ID}, a computer vision platform targeted at a non-specialist audience \cite{10.1145/3461702.3462627}. Their goal was to demonstrate disparate failure of facial recognition using a simple interaction mechanic: a ``match threshold'' slider, similar to the one I use in my platform. This was a powerful way to demonstrate how human policy choices lead to discriminatory outcomes. In the computer vision domain, there is yet another non-specialist tool, \textit{How Normal Am I}, which is a playful platform that manages to demonstrate algorithmic judgment (attractiveness, age, rating, etc.) \cite{hownormalami}. In the NLP domain, however, there exists no such project that aims at educating the public about linguistic bias. Therefore, I have drawn heavily from these two computer vision projects as references.

A noteworthy related project in the Hugging Face space, \textit{evaluate-measurement/toxicity}, which I stumbled later on during this research \cite{hf_toxicity}. Like my project, this space achieves a similar goal: quantification of toxicity in the input text, but it is based on a different hate speech classification model. My work can be seen as an extension, as my platform provides extended insights using other measurement metrics (e.g., identity attack, obscene, etc.) but its novelty lies in the focus on the \textbf{classification threshold} as the core interactive mechanic, which is essential for demonstrating how a biased score leads to a discriminatory outcome.

\section{Methodology}

This research was conducted on two fronts: first, building a quantitative data analysis pipeline to source and benchmark dialectal text samples, and second, developing an application framework to host the model and serve the interactive tool.

\subsection{Data Corpus Preprocessing}

For benchmarking purposes, I chose the 12 GB Twitter dataset \textit{twitteraae\_all} by Slanglab \cite{twitteraae}. This file contains millions of tweets, each with a corresponding set of probabilities from the original paper's demographic model \cite{blodgett-etal-2016-demographic}. My objective was to filter this massive dataset into two clean and commensurate corpora (10{,}000 samples each): one for African-American English (AAE) and the other for Standard American English (SAE). Each corpus contains high-confidence (>80\%) text classifications.

\begin{table}[ht]
\centering
\caption{Sample of \textit{twitteraae\_all} Dataset}
\begin{tabular}{lcc}
\hline
\textbf{Text} & \textbf{$p_{aa}$} & \textbf{$p_{white}$} \\
\hline
daqui a pouco vou ir KBLO..kkkkkk... & 0.099000 & 0.080000 \\
iLied\textbackslash nILied\textbackslash n\#iLied for a second i thought \#... & 0.446154 & 0.401538 \\
\textbackslash u0418\textbackslash u043d\textbackslash u0442\textbackslash u0435\textbackslash u0440\textbackslash u\textbackslash u0441\textbackslash u0... & 0.214444 & 0.183333 \\
Roll Tide Roll!!! \#2013 BCS National Champions & 0.000000 & 0.975000 \\
\@dustinpurcell All work and no play makes Jac... & 0.034545 & 0.909091 \\
\hline
\end{tabular}
\end{table}

The 12 GB dataset was not loaded into memory at once; rather, I read it in chunks of one million records at a time. I utilised the \textit{Pandas} dependency for this preprocessing. While parsing each chunk, I ignored all quoting characters and discarded malformed rows. I further sanitised the samples by removing URLs, social media handles (\texttt{@username}), special characters like hashes, newlines, and extra whitespace. In short, I ensured that the model analysed only the linguistic content.

\begin{table}[ht]
\centering
\caption{Sample of Sanitised AAE Dataset}
\begin{tabular}{lcc}
\hline
\textbf{Text} & \textbf{$p_{aa}$} & \textbf{$p_{white}$} \\
\hline
Tuh, who mad me or ol boy? & 0.875556 & 0.028889 \\
Man imissed a called from my bae hella mad… & 0.942000 & 0.000000 \\
Twink rude lol can't be calling ppl ugly that... & 0.815385 & 0.129231 \\
I did not mean to say dat & 0.957143 & 0.008571 \\
Smh & 0.900000 & 0.080000 \\
\hline
\end{tabular}
\end{table}

\begin{table}[ht]
\centering
\caption{Sample of Sanitised SAE Dataset}
\begin{tabular}{lcc}
\hline
\textbf{Text} & \textbf{$p_{aa}$} & \textbf{$p_{white}$} \\
\hline
Roll Tide Roll!!! 2013 BCS National Champions & 0.000000 & 0.975000 \\
Hey, you shouldn't insult dying cow… & 0.016000 & 0.822000 \\
I had a very laid back day. I wish I was in M... & 0.095484 & 0.830323 \\
I think I'm hungry... Not sure of what to eat... & 0.016250 & 0.904375 \\
I can finally receive videos on snapchat! I... & 0.014737 & 0.875789 \\
\hline
\end{tabular}
\end{table}

Despite the fact that users can enter their own text in the interactive tool, I prepared a list of example sentences grounded in robust empirical evidence. This choice is deliberate and justified for several reasons: first, this form of bias is well-documented in NLP research. Multiple studies have demonstrated that toxicity classifiers trained on standard, large-scale datasets are significantly more likely to flag text containing AAE lexical features as offensive, even when the content is benign \cite{field-etal-2021-survey}. Second, the social stakes are relatable. Most internet users have experience with content moderation and can understand the implications of being unfairly censored \cite{yang2023toxbusteringamechattoxicity}. Lastly, by presenting semantically identical sentences in two dialects, it becomes clear that the system is penalising a user not for \emph{what} they say, but for \emph{how} they say it. Example:

\begin{itemize}
\item \textbf{Dialectal Bias -- Zero Copula:}
  \begin{itemize}
    \item SAE: ``She is at the library studying.''
    \item AAE: ``She at the library studying.''
  \end{itemize}
\item \textbf{Dialectal Bias -- Double Negative:}
  \begin{itemize}
    \item SAE: ``I am not bothering anyone.''
    \item AAE: ``I ain't bothering nobody.''
  \end{itemize}
\end{itemize}

\subsection{Model Selection and API Framework}

I chose the \textit{unitary/toxic-bert} model \cite{toxicbert}, a \textit{google-bert/bert-base-uncased} model fine-tuned on the Jigsaw Toxic Comment Classification Challenge dataset \cite{bertbase, jigsaw}. I made this decision because this model is a widely-used, foundational, and publicly available classifier.

I developed a lightweight backend API in Python using Flask to serve this model. I specified the \textit{transformers} pipeline to use a sigmoid function, which ensures the model will return independent probabilities for all six of its output labels: toxicity, severe\_toxicity, obscene, threat, insult, and identity\_attack.

The final API exposes a single \texttt{/api/predict/} endpoint. This endpoint receives a JSON object with a text string, runs the full pipeline on it, and returns a JSON object containing the scores for all six labels.

\section{Demonstration}

To make the abstract findings of linguistic bias tangible, I developed an interactive pedagogical tool, the \textit{Dialectal Bias Analyser}. The tool is a self-contained application using a Flask Python backend to serve the model and an index.html file (using React and Babel.js) to render the user interface. I have explained the essential setup steps in depth in the project's GitHub repository \cite{github_analyser}.

\subsection{Web Interface}

\begin{figure}[ht]
\centering
\includegraphics[width=\columnwidth]{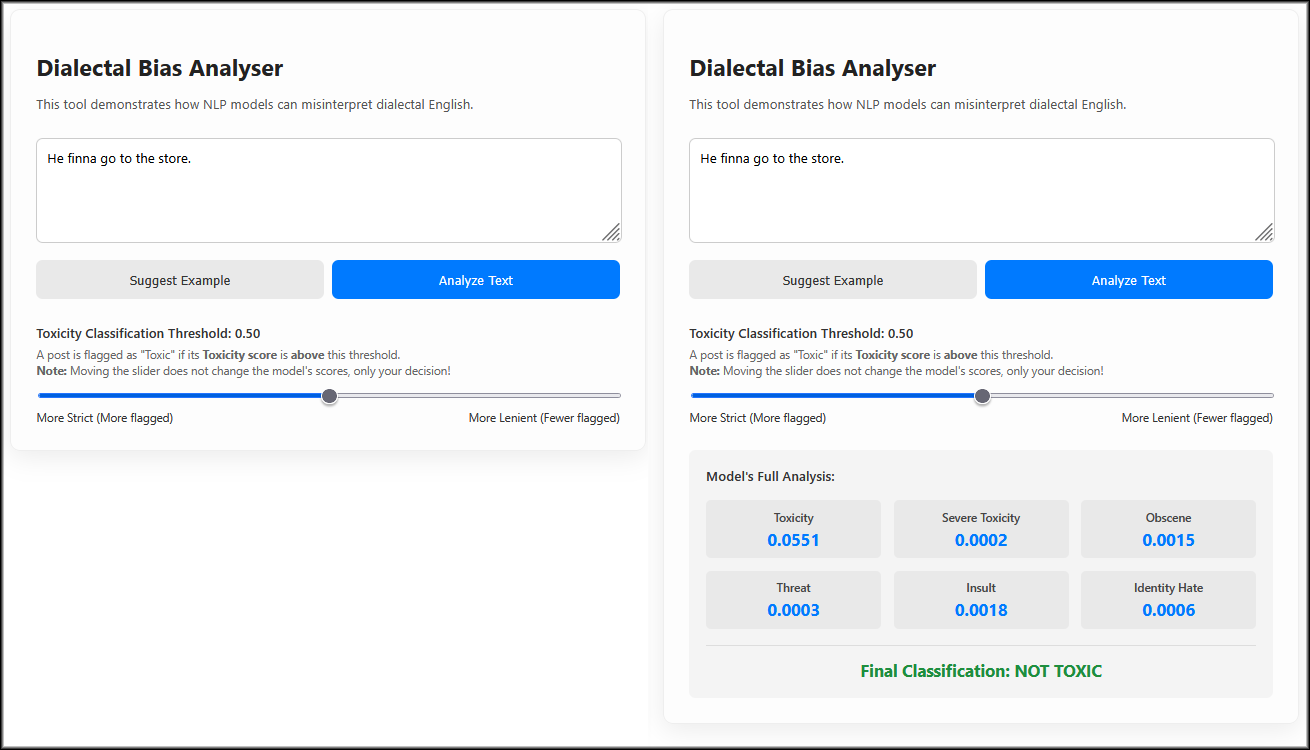}
\caption{Web interface of the Dialectal Bias Analyser.}
\label{fig:ui}
\end{figure}

The interface is intentionally minimal \cite{hf_space}. As depicted in Fig.~\ref{fig:ui}, a user is presented with a text area and two buttons: \textit{Suggest Example} and \textit{Analyse Text}. The \textit{Suggest Example} button populates the text area with a sentence from a pre-compiled list of ``minimal pairs'' I used in my own testing. The sentences range from dialectal bias to racial and gender bias to religion and identity bias. This serves as a basis for users to discover the bias.

Upon clicking \textit{Analyse Text}, the model's scores (aforementioned six output labels) are shown, based on the evaluation of the sentence in the text area.

\subsection{Threshold Slider: The ``Aha!'' Moment}

The central pedagogical feature of the tool is the \textit{Toxicity Classification Threshold} slider. This slider does not re-calculate the model's scores; rather, it acts as the human-set policy level. The final verdict (TOXIC/NOT TOXIC) displayed at the bottom is determined by a simple comparison: is the model's toxicity score greater than the user-controlled threshold?

\begin{figure}[ht]
\centering
\includegraphics[width=\columnwidth]{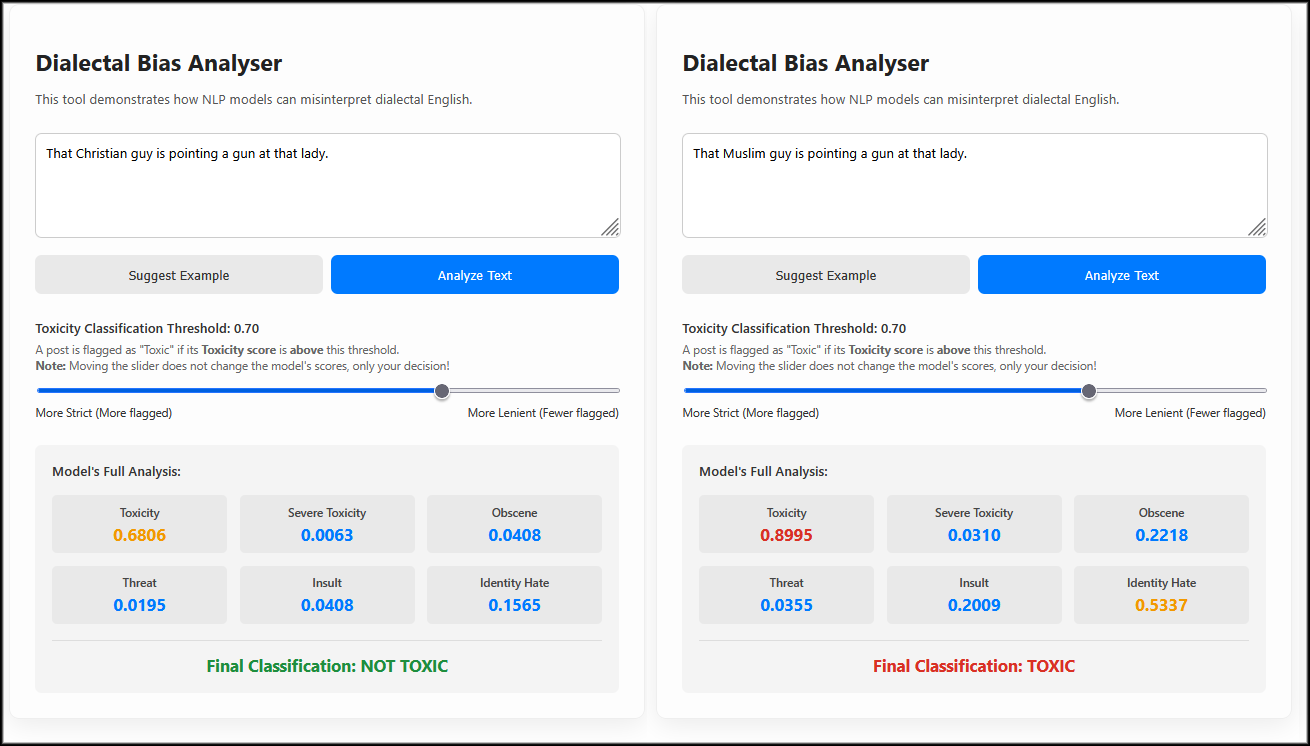}
\caption{Example of discriminatory outcome resulting from threshold adjustment.}
\label{fig:threshold}
\end{figure}

This mechanic is the ``Aha!'' moment. A user can analyse a sentence like ``That Christian guy is pointing a gun at that lady.'' and see it receives a low ``toxicity'' score (0.6806). They can then analyse ``That Muslim guy is pointing a gun at that lady.'' and see a higher score (0.8995). By setting the threshold slider to 0.70 (more lenient), they directly observe how a single, seemingly-neutral policy decision flags the post about the Muslim man but ignores the one about the Christian man (see Fig.~\ref{fig:threshold}). This interaction makes the concept of disparate impact tangible, demonstrating how a biased model, combined with a uniform policy, becomes an active mechanism of discrimination.

\section{Evaluation and Results}

To provide scientific validation for the tool's premise, I conducted a quantitative benchmark of the \textit{unitary/toxic-bert} model using a Jupyter Notebook (benchmark.ipynb), which can be found in the project's GitHub repository.

I used the 10{,}000-sample datasets discussed in Section~III-A and passed them through the \textit{transformers} pipeline to generate and store scores for all six measurement labels. In this section, I discuss my findings and their implications.

\subsection{Average Score Disparity}

The simplest analysis was to compare the mean scores obtained for the AAE and SAE corpora. 

\begin{table}[ht]
\centering
\caption{Average Scores by Dialect Group}
\label{tab:avg_scores}
\begin{tabular}{lcc}
\hline
\textbf{Label} & \textbf{AAE} & \textbf{SAE} \\
\hline
Toxicity & 0.279225 & 0.148181 \\
Severe toxicity & 0.030577 & 0.008457 \\
Obscene & 0.186225 & 0.076791 \\
Threat & 0.007578 & 0.005651 \\
Insult & 0.117351 & 0.042885 \\
Identity hate & 0.045805 & 0.005237 \\
\hline
\end{tabular}
\end{table}

The data in Table~\ref{tab:avg_scores} reveals a stark, quantitative disparity in the average scores assigned to each dialectal group. On average, the model scores AAE text as \textbf{1.8 times more toxic} and \textbf{8.8 times higher for \textit{identity hate}}. This is a clear systematic bias from a computer model that the average population supposes to be neutral, fair, and just: ``If the AI has said this, then it must be true.''

\subsection{Score Distribution Analysis}

To ensure these averages were not skewed by a few outliers, I plotted the full distribution of ``toxicity'' scores for both groups using a box plot.

\begin{figure}[ht]
\centering
\includegraphics[width=\columnwidth]{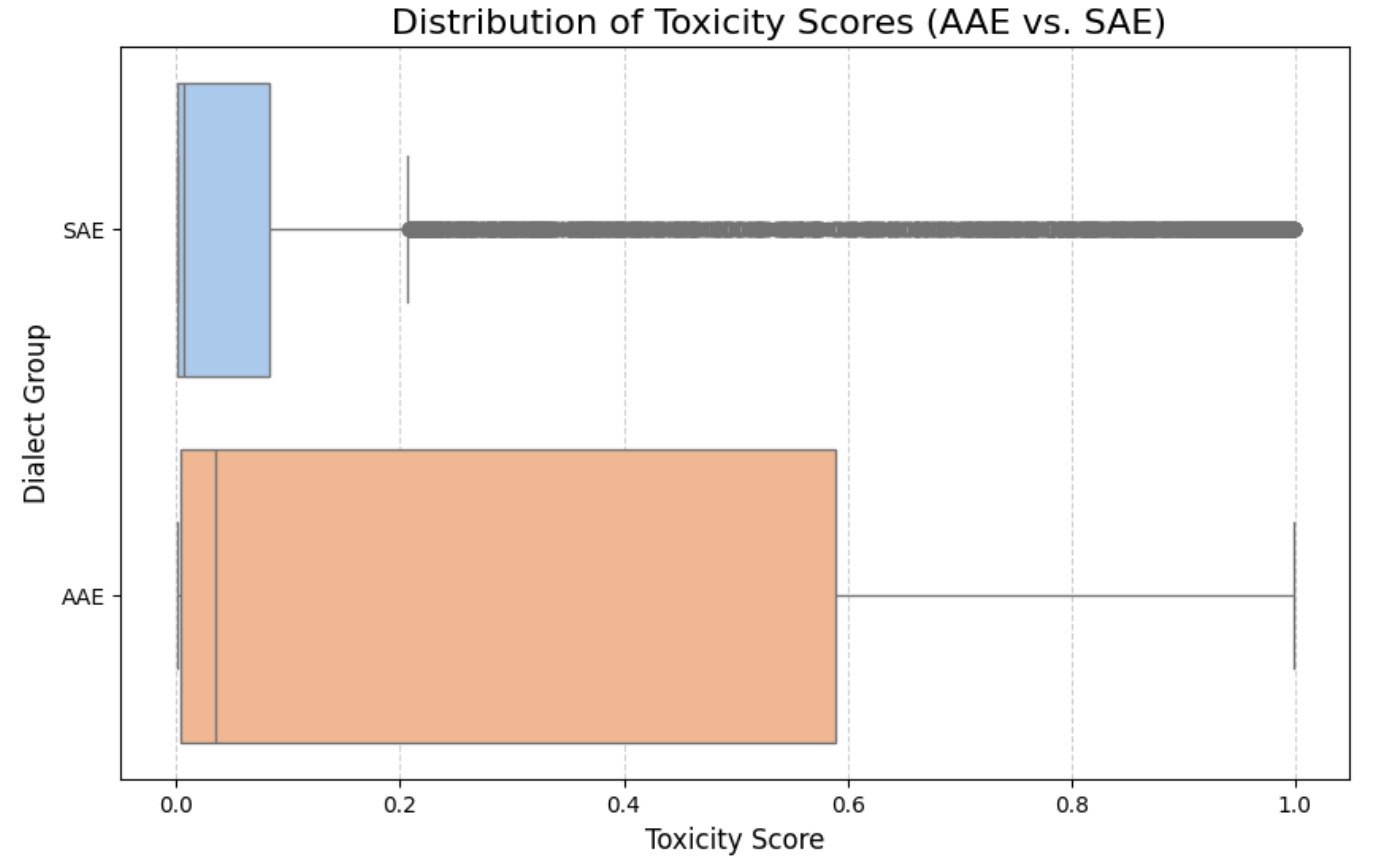}
\caption{Distribution of toxicity scores for AAE and SAE text (box plot).}
\label{fig:boxplot}
\end{figure}

\begin{figure}[ht]
\centering
\includegraphics[width=\columnwidth]{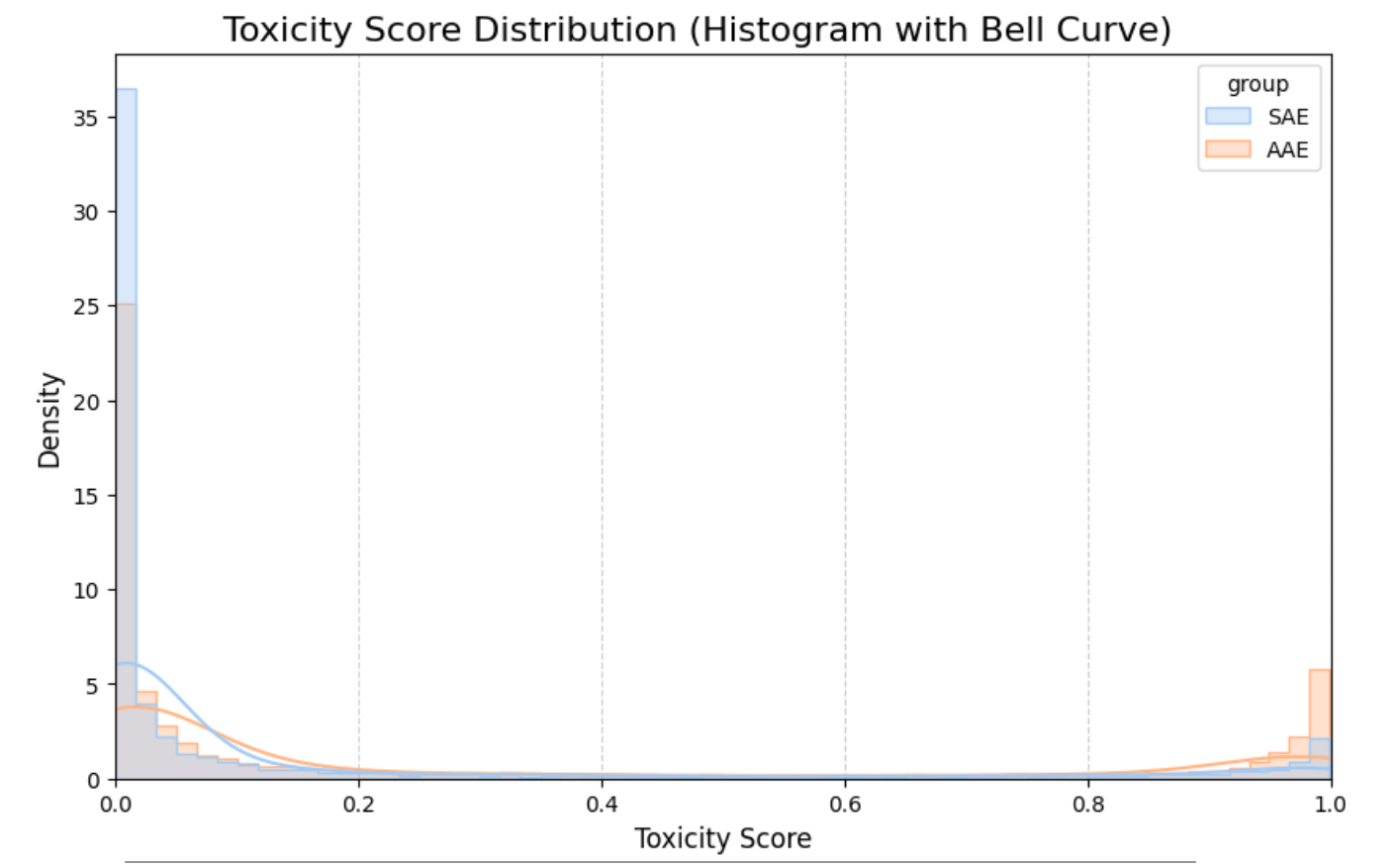}
\caption{Distribution of toxicity scores for AAE and SAE text (histogram).}
\label{fig:histogram}
\end{figure}

Figure~\ref{fig:boxplot} visually confirms the systematic nature of the bias. The box plot for SAE (Blue) is compressed near zero, meaning the vast majority of SAE posts received a toxicity score very close to 0. Any abnormally high scores are rendered as outliers. In contrast, for AAE (Orange), even higher toxicity scores are treated as part of the normal, expected results. The AAE box is much wider and shifted to the right, with a median at around 0.05. Fig.~\ref{fig:histogram} complements this finding, showing the SAE curve has a massive spike at 0 (confirming most posts are scored as non-toxic), while the AAE curve is much flatter and spread out.

\subsection{False Positive Rate (FPR) Analysis}

\begin{figure}[ht]
\centering
\includegraphics[width=\columnwidth]{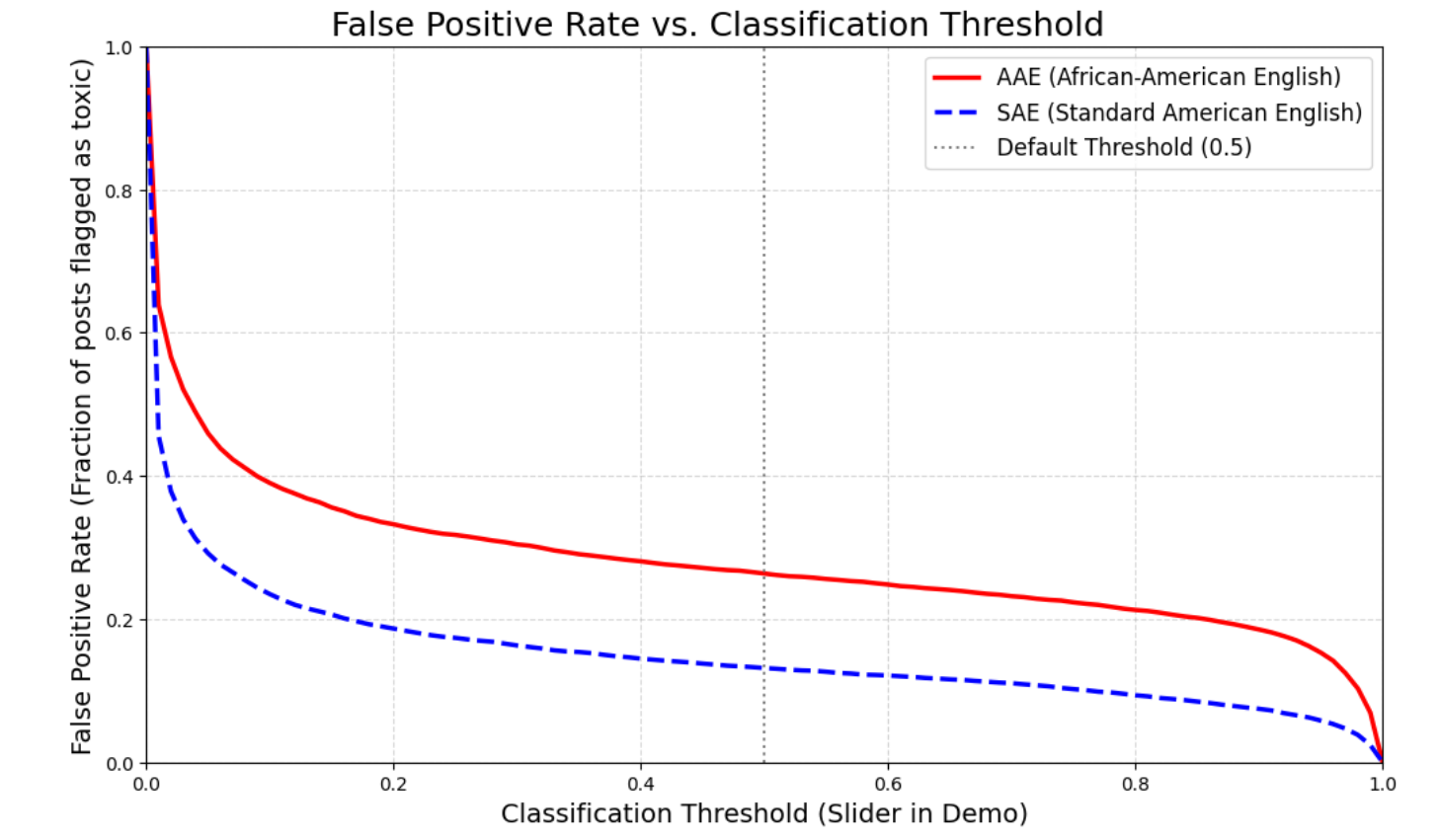}
\caption{False positive rate as a function of classification threshold.}
\label{fig:fpr}
\end{figure}

The most critical finding is displayed in Fig.~\ref{fig:fpr}. The False Positive Rate (FPR) is the rate at which benign text is incorrectly flagged as ``toxic''. This corresponds to the ``threshold'' slider in the interactive tool. As the classification threshold shifts from 0.0 (stricter, more flags) to 1.0 (more lenient, fewer flags), the FPR score changes for both AAE and SAE posts. At every point, except the absolute extremes, the AAE (Red) line is consistently higher than the SAE (Blue) line. This proves that at any given sensitivity level, AAE text is more likely to be incorrectly flagged as toxic. This means that no matter what threshold the human operator sets, the AAE community will be more prone to automated disparity.

\section{Discussion}

The results from the quantitative benchmark are unambiguous. They are not a series of random errors but a reproducible pattern. The model exhibits a clear, systematic bias against one group. Why does this happen? Computers do not have an inner perspective towards any group; they are not supposed to be biased. How can an indiscriminatory system exhibit so much discrimination? Its flaw lies in its training.

Unlike humans, an algorithmic program is not ``racist''; it is a pattern-recognition engine. Even advanced Machine Learning (ML) and Artificial Intelligence (AI) systems, perhaps more accurately described as Quasi-Intelligence (QI), operate based on complex statistical analysis of data. This is just another form of advanced pattern recognition, not a human-like conscience or awareness.

Models like \textit{unitary/toxic-bert} are trained on massive internet datasets (e.g., comments from Wikipedia, Reddit, etc.). As a non-native English speaker, I learned Standard American English through my school curriculum and learned to write content for the internet in SAE. I assume other non-native speakers learn SAE the same way. Hence, data on the internet is overwhelmingly written in SAE. Furthermore, proofreading tools are developed to default to SAE corrections. Therefore, in a digital society, SAE is the norm, and other dialects like African-American English, Irish English, or Australian English are underrepresented. Because the dialectal features of these languages are ``uncommon'' in the training data, the model learns to associate these ``abnormal'' linguistic patterns with other ``abnormal'' patterns it was trained to detect, namely, profanity and toxicity. This is called \textbf{False Association} through \textbf{Training Data Imbalance}.

By introducing a sensitivity slider, we are not eliminating the bias ingrained in the model; we are simply placing control in a human’s hands to justify its operationalisation. There is no ``fair'' threshold. A content screening team might set a ``neutral'' policy (e.g., 0.5), believing it to be an objective choice, but at any sensitivity level, one group is silenced more than the other for saying the exact same thing in a different dialect. This isn’t just a technical flaw; it is a form of automated linguistic discrimination. By systematically flagging one demographic group, these tools create a digital space that is less welcoming and more hostile towards speakers of non-standard dialects, reinforcing existing social inequalities.

\section{Limitations and Future Work}

This study provides a clear demonstration of bias, and I have made some bold claims based on the study outcomes. However, it is important to acknowledge its limitations.

\begin{itemize}
    \item This study focused on a single, foundational model. As noted on the model's own documentation, the \textit{unitaryai/detoxify} library is now the recommended successor. I have not yet verified if this newer model has mitigated the bias or simply shifted it.
    
    \item Since I sourced the ``ground truth'' for classifying text from the TwitterAAE dataset, which itself used a model for classification, this is a model-on-model analysis, not a comparison against human-annotated linguistic data.

    \item For the False Positive Rate analysis, I assumed that all 20{,}000 samples were benign. This is effective for demonstrating a general disparity, but it is not a pure measure of FPR, as some tweets could genuinely be toxic.
\end{itemize}

These limitations pave a clear path for future research.

\begin{itemize}
    \item First and foremost, it is critical to re-run this exact benchmark methodology on newer and different models, including \textit{unitaryai/detoxify} and models from Jigsaw's more recent challenges. This would reveal whether the problem is being solved by the industry or if it is still persisting.
    
    \item The same methodology could be applied to investigate more specific biases, such as those related to gender or religion, by deriving a dataset that focuses explicitly on those demographic contrasts.

    \item A more robust human-in-the-loop validation involving a smaller, human-annotated dataset, where texts are labelled for both dialect and ground-truth toxicity by linguists, would further affirm this study's results.

    \item The interactive tool could be expanded to allow users to select and compare multiple models side-by-side, along with displaying their result score in a chart, creating an even more powerful pedagogical dashboard.
\end{itemize}

\section{Conclusion}

I successfully built and validated a dual-pronged system: a quantitative benchmark that proves the existence of dialectal bias, and an interactive tool that demonstrates the mechanism of its harm. The benchmark results are clear: the \textit{unitary/toxic-bert} model, a foundational tool for content moderation, scores African-American English as \textbf{1.8 times more toxic} and \textbf{8.8 times more likely to contain ``identity hate''} than Standard American English.

The more significant contribution, however, is the \textit{Dialectal Bias Analyser} tool itself. By giving users control of the ``sensitivity threshold,'' the tool moves the conversation beyond just a biased score and reveals the true harm: a biased outcome. It provides a tangible ``Aha!'' moment, proving that a seemingly neutral, human-set policy is the lever that operationalizes discrimination. By making this mechanism visible and interactive, this work provides a public-facing tool to foster the critical AI literacy needed to question these ``black box'' systems and advocate for more equitable technology.

This paper is not the first to find this bias, and I am sure it will not be the last. This is a widely documented problem in the AI Ethics and Fairness community. However, this work contributes more than just another statistical report; it provides a direct bridge between that expert knowledge and the public's understanding.

\bibliographystyle{IEEEtran}
\bibliography{references}

\end{document}